\documentclass[conference]{IEEEtran}
\IEEEoverridecommandlockouts
\usepackage{cite}
\usepackage{amsmath,amssymb,amsfonts}
\usepackage{algorithmic}
\usepackage{graphicx}
\usepackage{textcomp}
\usepackage{xcolor}
\usepackage{multirow}
\usepackage{adjustbox}
\usepackage{booktabs}
\usepackage{hyperref}
\usepackage[flushleft]{threeparttable}
\def\BibTeX{{\rm B\kern-.05em{\sc i\kern-.025em b}\kern-.08em
    T\kern-.1667em\lower.7ex\hbox{E}\kern-.125emX}}
\begin{document}

\makeatletter
\newcommand{\linebreakand}{%
  \end{@IEEEauthorhalign} 
  \hfill\mbox{}\par
  \mbox{}\hfill\begin{@IEEEauthorhalign}
}
\makeatother

\title{GraSSNet: Graph Soft Sensing Neural Networks\\
\thanks{\textsuperscript{$\dagger$} Work performed while at Seagate. \textsuperscript{*} Corresponding Author.}
}

\author{

\IEEEauthorblockN{Yu Huang\textsuperscript{1,2,$\dagger$}, Chao Zhang\textsuperscript{1,3}, Jaswanth Yella\textsuperscript{1,4}, Sergei Petrov\textsuperscript{1,5}, Xiaoye Qian\textsuperscript{1,6}}
\IEEEauthorblockA{\textsuperscript{1}Seagate Technology, \textsuperscript{2}Florida Atlantic University, \textsuperscript{3}University of Chicago, \textsuperscript{4}University of Cincinnati,\\\textsuperscript{5}Stanford University, \textsuperscript{6}Case Western Reserve University\\
Email: \{yu.1.huang, chao.1.zhang, jaswanth.k.yella, sergei.petrov, xiaoye.qian\}@seagate.com}

\linebreakand 
\IEEEauthorblockN{Yufei Tang\textsuperscript{2}, Xingquan Zhu\textsuperscript{2}}
\IEEEauthorblockA{Florida Atlantic University, FL, USA\\
Email: \{tangy, xzhu3\}@fau.edu}

\and
\IEEEauthorblockN{Sthitie Bom\textsuperscript{1,*}}
\IEEEauthorblockA{Seagate Technology, MN, USA\\
Email: sthitie.e.bom@seagate.com}}

\IEEEoverridecommandlockouts
\IEEEpubid{\makebox[\columnwidth]{This paper has been accepted by 2021 IEEE International Conference on Big Data \hfill} \hspace{\columnsep}\makebox[\columnwidth]{ }}

\maketitle

\newcommand{\modelname}{GraSSNet}

\begin{abstract}
In the era of big data, data-driven based classification has become an essential method in smart manufacturing to guide production and optimize inspection. The industrial data obtained in practice is usually time-series data collected by soft sensors, which are highly nonlinear, nonstationary, imbalanced, and noisy. Most existing soft-sensing machine learning models focus on capturing either intra-series temporal dependencies or pre-defined inter-series correlations, while ignoring the correlation between labels as each instance is associated with multiple labels simultaneously. In this paper, we propose a novel graph based soft-sensing neural network (\text{\modelname}) for multivariate time-series classification of noisy and highly-imbalanced soft-sensing data. The proposed \text{\modelname} is able to 1) capture the inter-series and intra-series dependencies jointly in the spectral domain; 2) exploit the label correlations by superimposing label graph that built from statistical co-occurrence information; 3) learn features with attention mechanism from both textual and numerical domain; and 4) leverage unlabeled data and mitigate data imbalance by semi-supervised learning. Comparative studies with other commonly used classifiers are carried out on Seagate soft sensing data, and the experimental results validate the competitive performance of our proposed method.
\end{abstract}


\section{Introduction}
Industry 4.0, which encompasses the Internet of Things (IoT)  and smart manufacturing, marries physical production and operations with smart digital technology, machine learning, and big data to create a more holistic and better connected ecosystem for industries that focus on high-tech manufacturing \cite{zhou2015industry}. Due to the increase in complexity and cost, the manufacturing industry, such as semiconductor manufacturing \cite{fan2020defective}, is becoming more and more complicated. To improve the production efficiency and quality control, the direct, fast, and accurate measurement and analysis/inspection of \textit{key quality indicators} (KQIs) are in rising demand. In response, soft-sensing models have been developed to estimate/predict KQIs expediently during the past decades, which is usually formulated as a mathematical model with \textit{easy-to-measure} auxiliary variables as inputs and \textit{hard-to-measure} key indicators as outputs \cite{sun2021survey}. While soft-sensing models are of the process monitoring and diagnosis tasks, this paper mainly focus on diagnostic applications, i.e. multi-label multivariate time series classification problem. 

To establish a soft-sensing diagnostic model, two major categories of methodologies are widely adopted, mechanism/knowledge-based and data-driven-based method \cite{venkatasubramanian2003review}. The former requires expert knowledge (or a wealth of experience) of detailed and accurate mechanism of the manufacturing process, which is hard to meet (acquire) with the increasing complexity of the industrial processes. In contrast, data-driven-based method (esp., deep learning models) is `\textit{winning}' in the field of soft sensing technology. The improvements in data availability and computational scale have been the dominant driving force behind data-driven modeling.

With the rapid development of smart digital technology and the wide use of the distributed control systems \cite{geng2021novel}, more and more complex and ever-evolving process data are generated and stored in huge amounts, as monitoring sensors are increasingly installed in factories to measure real-time process status (e.g., temperature, pressure, etc.). Such data have the attributes of high nonlinearity, high-dimension, imbalance, and noise. How to make full use of industrial big data to effectively improve diagnostic performance, as well as avoid complicated feature engineering and learn abstract representation automatically, have become a challenging problem in developing cost-effective and scalable methods.

Traditional data-driven soft-sensing models like the kernel principal component analysis \cite{dai2020temperature}, support vector machine \cite{wang2019two} and artificial neural networks \cite{xuefeng2010hybrid} have been introduced for fault classification in industrial processes. However, such models show limitations in handling multi-mode, high-dimensional, noisy, and imbalanced data. Recently, deep neural networks have achieved breakthrough results and exhibit stronger capabilities in learning and representation over traditional methods, such as stacked auto-encoder \cite{yuan2020stacked,zhang2021auto} and convolutional neural networks \cite{lei2020semi}. Despite a proliferation of research that applies deep learning approaches to soft sensing, there are several aspects that need to be further investigated, considering the multi-label multivariate time series classification scenario. 

The industrial data obtained in practice is usually multivariate time-series data collected by soft sensors. It is challenging since soft sensing models need to consider both intra-series temporal correlations and inter-series correlations jointly. Deep learning models, such as long short-term memory \cite{yuan2020deep} and temporal convolution networks \cite{koh2021deep}, have achieved promising results in temporal modeling. However, most of them ignore modeling the correlations among multiple time-series. Recently, some novel works \cite{li2017diffusion} tried to learn both correlations by stacking graph convolution neural networks (GCN) \cite{kipf2016semi} to temporal modules, where GCN was designed to capture inter-series relationships explicitly based on pre-defined sensor topology.

Modeling the label dependencies is important in soft sensing since the collected data are usually associated with multiple labels. In physical world, some combinations of labels are almost impossible to appear, while some are coincident with high possibility. Many previous classifiers are essentially limited, ignoring the complex topology between labels. This vitalizes research in exploring the label correlations, including graph learning models \cite{shi2019mlne,chen2019multi}, recurrent neural networks-based model \cite{wang2016cnn}, and attention mechanisms \cite{wang2017multi}. Graph-based models have been proven to be more effective in modeling label correlation \cite{chen2019multi,shi2020multi}. However, to the best of our knowledge, the label correlation is mostly explored in image classification, under-explored in soft sensing.

In this paper, a novel graph based soft-sensing neural network (\text{\modelname}) model is proposed for complex industrial process inspection by classifying KQIs (labels) given multivariate time series. This paper presents the first empirical study of wafer inspection challenge addressed by the competition that we organized in frame of the IEEE BigData 2021 Cup. in \textit{Soft Sensing at Scale - Seagate}\footnote{https://github.com/Seagate/BigDataChallenge}. We formulate the problem as multi-label classification, since diagnostic KQIs are not mutually exclusive. The main contributions of this paper are:
\begin{enumerate}
    \item \text{\modelname} is proposed to capture the intra-series temporal patterns and inter-series sensor correlations jointly in the spectral domain, where spectral representations hold clearer patterns and can be classified more effectively. \text{\modelname} enables a data-driven construction of dependency graphs for different time series without pre-defined topologies.
    \item To make full use of various types of data, the dependency between textual information and numerical time series data is modeled through an attention mechanism. 
    \item Graph attention networks are used to propagate information between multiple labels to explicitly model the label dependencies, where the label correlation matrix is defined based on their co-occurrence patterns.
    \item Extreme negative-positive imbalance and high unlabeled rate -- which are typical challenges in soft sensing -- are explicitly addressed by training the model through a joint loss function.
\end{enumerate}

\section{Related work}
Our research sits at the intersection of soft sensing, multi-label classification, and imbalance learning. Three mature fields, each with a long history and rich body of research. While we cannot do justice to all three, we highlight the most relevant works below.

\subsection{Deep Learning in Soft Sensing}
Soft sensors are widely constructed in factories to realize process monitoring, quality prediction, and other important industrial applications \cite{sun2021survey,qian2020wearable}. Recently, improvements in big data and computational scale have driven a proliferation of research that apply deep learning approaches to soft sensors. Autoencoders are usually adopted to extract feature representations \cite{huang2021prognostics} and handling missing data issues \cite{huang2020reliable, guo2020output}. Convolutional neural networks (CNNs) are suited for processing grid data in capturing local dynamic characteristics \cite{wang2019dynamic} or processing signals in the frequency domain \cite{yuan2020soft,wei2015soft,qian2019smart}. Recurrent neural networks (RNNs) and their variants LSTMs and GRUs based soft sensors were developed to estimate variables with strong temporal patterns \cite{zhang2019automatic}, and to cope with strong nonlinearity and dynamics of the process \cite{ke2017soft}. With various machine learning based soft sensing model proposed for different aspects, there is still much to be done to better apply the advanced methods in the soft sensing domain, especially to meet the ever-changing demands in practical industrial processes.

\subsection{Multi-Label Classification}
Multi-label classification is a fundamental and practical task in machine learning, where the aim is to predict a set of labels related to a sample. In most multi-label tasks, labels are treated in isolation and converted into a set of binary classification problems to predict whether each label of interest presents or not. Deep CNNs \cite{huang2017densely,zheng2016exploiting,chen2021time}, RNNs/LSTMs \cite{karim2019multivariate,lipton2015learning}, or hybrid models \cite{guo2020multivariate} are widely used and have achieved promising results. However, a key characteristic that distinguishes the multi-label from multi-class classification is the combinatorics of the output space \cite{chen2019multi}. Many researchers attempted to regularize the prediction space by capturing label dependencies. Notable success was reported by explicitly modeling label dependencies via graph model \cite{chen2019multi,li2014multi,tan2015learning, durand2019learning} or word embedding based on knowledge priors \cite{wang2020multi,chen2019multi}, while some work implicitly modeled the label correlations via attention mechanisms \cite{wang2017multi,you2020cross}.

\begin{figure}[t]
\centering
\includegraphics[width=\linewidth]{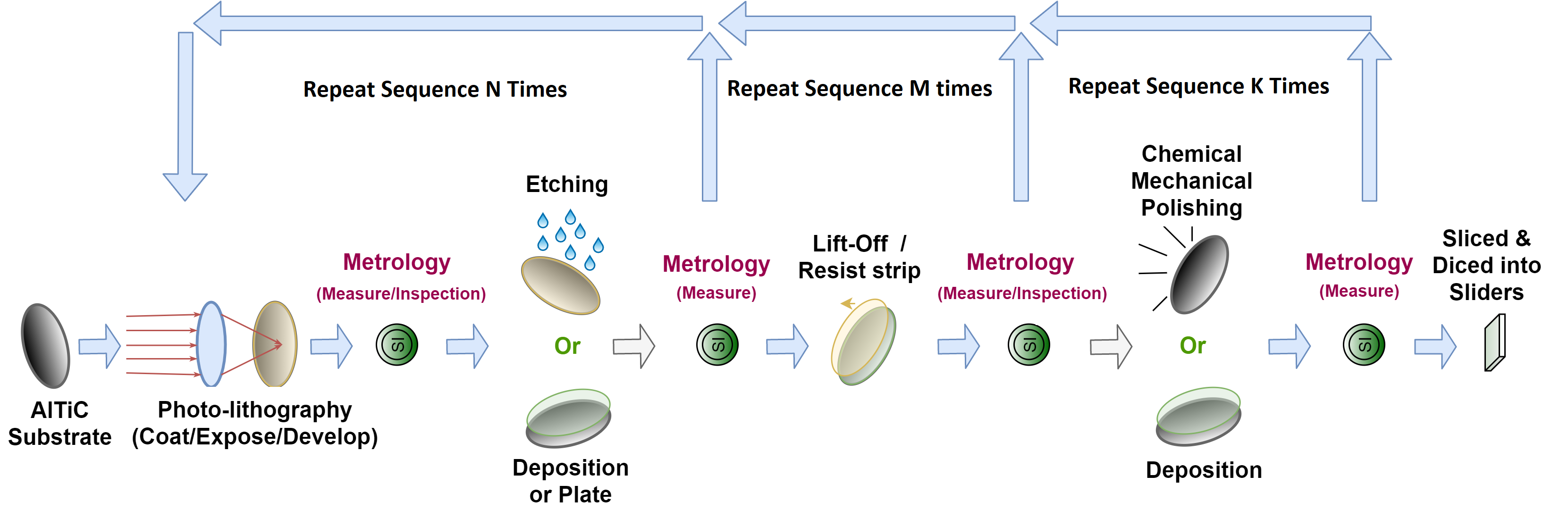}
\caption{Illustration of the wafer manufacturing process. Each wafer goes through multiple processing stages with corresponding meteorology where a few quality control measurements are performed. Figure from \url{https://github.com/Seagate/softsensing\_data}.}
\label{fig_wafer}
\end{figure}

\subsection{Imbalanced Classification}
Another key characteristic of multi-label classification is the inherent positive-negative imbalance. Most samples may contain only a small fraction of the candidate labels, implying that the number of positive samples per category will be much lower than the number of negative samples. Some re-sampling methods \cite{oksuz2019imbalance} were proposed by only selecting a more balanced subset. However, such methods are not suitable for handling imbalanced multi-label classification, since each sample contains many labels and re-sampling cannot change the distribution of only a specific label \cite{ben2020asymmetric}. Another common solution is to adopt a modified loss function \cite{wu2020distribution} to train on all examples without sampling and without easy negatives overwhelming the loss and computed gradients. For example, focal loss \cite{lin2017focal,chen2019learning,chen2019multi} puts focus on hard samples while down-weighting easy samples, by decaying the loss as the label’s confidence increases. More recently, asymmetric loss \cite{ben2020asymmetric} focuses on hard negatives while maintaining the contribution of positive samples by decoupling the modulations of the positive and negative samples and assigning them with different exponential decay factors. It also shifts the probabilities of negative samples to completely discard very easy negatives.

\section{Problem Definition}
\subsection{Soft Sensing at Scale - Seagate}
The manufacturing process of wafers is complicated and time-consuming. As shown in Fig. \ref{fig_wafer},  each wafer undergoes several permutations of the processing stages, including metal deposition, dielectric deposition, etching, electroplating, planarization, and lithograph \cite{quirk2001semiconductor}. Due to the multiple complicated processing stages, it is difficult to guarantee manufacturing stability at any time, which limits the quality control in actual industrial production. To improve the predictability of qualified product yield, a large sensor network is installed in the manufacturing line to monitor the wafer quality. At each processing stage, multiple critical sensor records are collected. The engineers at Seagate inspect these records and attest to the quality of the wafer based on some internal heuristic threshold values for each KQIs. However, the collected sensor data have the characteristics of high nonlinearity, dynamics, and noise, requiring a high labor force to handle. An efficient data-driven model is in demand to predict the inspection results (pass/fail of multiple binary indicators) based on the multivariate time-series sensor data. The above problem was presented in the big data challenge - \textit{Soft Sensing at Scale - Seagate} - that we organized in frame of the IEEE BigData Cup 2021. The dataset released has 11 inspection KQIs (labels), with characteristics of high negative-positive rate, high unlabeled rate, and irregular time length. Statistic details of the dataset refer to Section \ref{section:data}.

\subsection{Multi-Label Classification}
Let $\textbf{L}=\{l_1, \cdots, l_N\}$ be a finite set of binary class labels and $\mathbb{X}$ denote an input space. Suppose every instance $x\in\mathbb{X}$, where $x\in\mathbb{R}^d$, is associated with a subset of labels $\overline{\textbf{L}}\subset\textbf{L}$, i.e., the set of relevant labels. The complement set of $\overline{\textbf{L}}$ is called the irrelevant set. Therefore, $D = \{(x_1, \overline{\textbf{L}}_1), (x_2, \overline{\textbf{L}}_2), \cdots, (x_n, \overline{\textbf{L}}_n)\}$ is a finite set of training instances that are assumed to be randomly drawn from an unknown distribution. The objective is to train a multi-label classifier $f: \mathbb{X} \rightarrow 2^{\textbf{L}}$ that best approximates the training data and generalizes well to the samples in the test data.

\begin{figure*}
    \centering
    \includegraphics[width=0.95\linewidth]{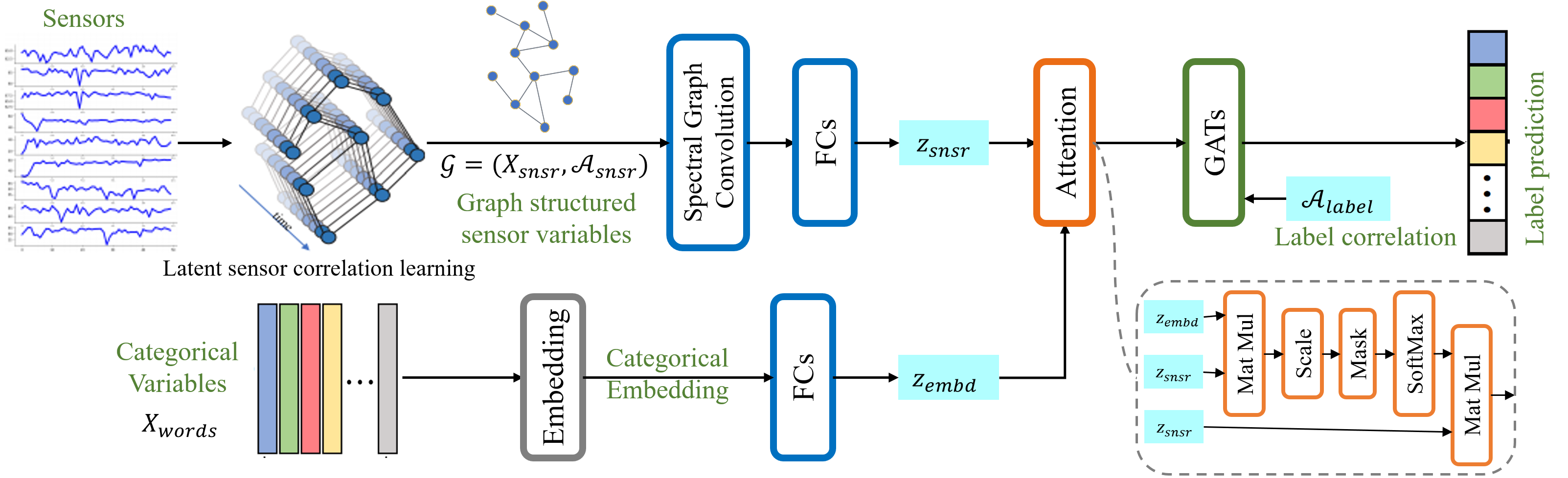}
    \caption{Overall architecture of the proposed Soft-sensing Graph Neural Network (\text{\modelname}).}
    \label{fig:ssgnn}
\end{figure*}

\begin{figure}
    \centering
    \includegraphics[width=\linewidth]{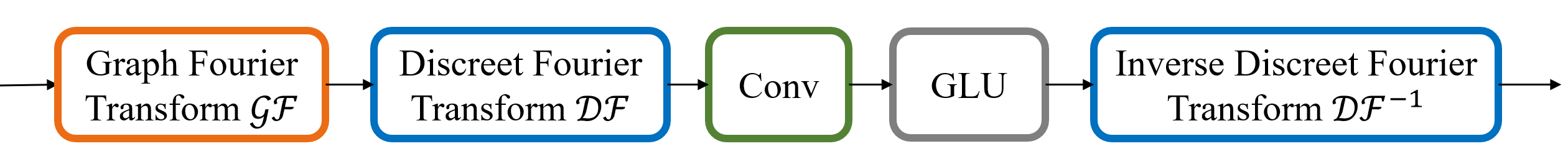}
    \caption{Spectral Graph Convolution Module.}
    \label{fig:spectralGCNblock}
\end{figure}

\section{Preliminaries}
Graph convolutional networks (GCNs) are a generalization of well-established convolutional neural networks to non-Euclidean graph-structured data. It can leverage graph topology to aggregate node information from the neighborhood in a convolutional fashion, following the widely-adopted GCN version proposed by \cite{kipf2016variational}, which is a spectral-based graph convolution design with spatial localization meaning. Assume we have a graph $\mathcal{G}$ with $N$ nodes, whose topology is represented by an adjacency matrix $\mathcal{A} \in \mathbb{R}^{N\times N}$. By projecting the graph to an orthonormal space where the bases are constructed by eigenvectors of the normalized graph Laplacian, the corresponding normalized graph Laplacian is defined as,
\begin{equation}
    \L=I_N-D^{-\frac{1}{2}}\mathcal{A}D^{\frac{1}{2}}=U\Lambda U^T
\end{equation}
where $D\in\mathbb{R}^{N\times N}$ is the diagonal degree matrix with $D_{ii}=\sum_j \mathcal{A}_{ij}$, $I_N\in \mathbb{R}^{N\times N}$ is the identity matrix, and $U$ and $\Lambda$ are eigenvectors and diagonal matrix of eigenvalues corresponding to Laplacian matrix $\mathcal{L}$, respectively. The spectral convolution on the graph is,
\begin{equation}
    \mathcal{G}_{\theta}\star x=U\mathcal{G}_{\theta}U^T x
\end{equation}
where $x\in\mathbb{R}^N$ is a graph feature vector and $\mathcal{G}_{\theta}$ is the graph convolution kernel. The intuitive explanation is that a graph Fourier transform is first applied on graph features by $U^T x$, and then multiplied by the convolution kernel $\mathcal{G}_{\theta}$, and finally a inverse Fourier transform is performed by multiplying it with $U$. Therefore, the operators of Graph Fourier Transform (GFT) and Inverse Graph Fourier Transform (IGFT) are defined as,
\begin{equation}
    \mathcal{GF}(x) = U^T x = \hat{x}; \; \mathcal{GF}^{-1}(\hat{x}) = U \hat{x}
\end{equation}

Next, we regard the convolution kernel as a polynomial function $\mathcal{G}_{\theta}(\Lambda)$ of the diagonal eigenvalue matrix $\Lambda$, so that the convolution becomes,
\begin{equation}
\begin{split}
    \mathcal{G}_{\theta} \star x & = U\mathcal{G}_{\theta}(\Lambda)U^T x\\
    &=U\left(\sum_{k=0}^K \theta_k\Lambda\right)U^T x = \sum_{k=0}^K \theta_k\Lambda^k x
\end{split}
\end{equation}
where $K$ is the chosen order of polynomial approximation, and $\theta$ are trainable parameters. To further improve the computational efficiency, we approximate $\mathcal{G}_{\theta}(\Lambda)$ by its Chebyshev polynomials and set the order of approximation $K=1$. The Chebyshev polynomials are recursively defined as $T_k(x) = 2xT_{k-1}(x)-T_{k-2}(x)$, with $T_0(x)=1$ and $T_1(x)=x$. To meet the requirement of Chebyshev polynomials, we normalize the eigenvalues as $\hat{\Lambda}={2}/(\lambda_{max}\Lambda-I_N)$ to make them lie within $[-1,1]$. $\lambda_{max}$ denotes the largest eigenvalue of $\L$, which is assumed to be 2. After a few derivation steps, the graph convolution becomes:
\begin{equation}
\begin{split}
    \mathcal{G}_{\theta} \star x & = {\theta_0}'x+{\theta_1}'(\L-I_N)x \\
    & = \theta(I_N+D^{-\frac{1}{2}}\mathcal{A}D^{-\frac{1}{2}})x = \theta\Tilde{D}^{-\frac{1}{2}}\Tilde{\mathcal{A}}\Tilde{D}^{-\frac{1}{2}}x
\end{split}
\end{equation}
with a single parameter $\theta={\theta_0}'=-{\theta_1}'$, $\Tilde{\mathcal{A}}=\mathcal{A}+I_N$ and $\Tilde{D}_{ii}=\sum_j\Tilde{\mathcal{A}}_{ij}$. Though derived from the spectral domain, the graph convolution above is considered to have a clear meaning of spatial localization \cite{zhang2019graph}. It is essentially equivalent to aggregating node representations from their direct neighborhood each time. Finally, the graph convolution cell can be defined as:
\begin{equation}
    \textup{Y}=\sigma \left( \Tilde{D}^{-\frac{1}{2}}\Tilde{\mathcal{A}}\Tilde{D}^{-\frac{1}{2}}\textup{X}\Theta \right)
\end{equation}
where $\textup{X}$ is the input, $\Theta$ is the trainable parameter matrix, and $\sigma$ is the sigmoid activation function.

\section{Methodology}
\subsection{Overall Framework}
We propose a Graph based Soft-sensing Neural Network (\text{\modelname}) as a scalable solution for multivariate time-series classification in the soft sensing. The overall architecture of \text{\modelname} is illustrated in Fig. \ref{fig:ssgnn}. It has two branches processing two types of data stream, i.e., numerical sensor records $X_{snsr}$ and textual information $X_{words}$.

In the first branch, the multivariate time-series input $X_{snsr}$ is first fed into a latent correlation layer to automatically infer the graph structure (i.e., the soft sensors network topology) and its associated weighted adjacent matrix $\mathcal{A}_{snsr}$. Next, the graph $G = (X_{snsr};\mathcal{A}_{snsr})$ serves as input to the spectral graph convolution module that is designed to model structural and temporal dependencies inside multivariate time-series jointly in the spectral domain (as visualized in Fig. \ref{fig:spectralGCNblock}). After spectral graph convolution module, feature representations on frequency basis are obtained by decomposing each individual time-series. Then, an output layer composed of fully-connected (FC) sub-layers is added to generate sensor feature representations with lower dimension. In the second branch, an embedding layer is used to encode lexical semantics, following with a FC sub-layers to learn a textual feature representations.

To prioritize and leverage the important distinctive features in sensor and textual feature representations, we introduce an attention mechanism to learn a feature fusion to boost performance. Finally, a graph attention network module is attached to capture the label correlations for multi-label classification and obtain the final predicted scores.

\subsection{Latent Graph Learning Module}\label{model2}
Graph neural network based approach requires a graph structure. It can be artificially constructed by human knowledge, such as using thresholded Gaussian kernel \cite{shuman2013emerging} to compute the pairwise road network distances between distributed sensors in traffic forecasting. However, sometimes we do not have a pre-defined graph structure as prior, such as in this paper, the sensor network topology is unknown. To tackle this problem, we leverage the self-attention mechanism to exploit the correlations between sensors, i.e. learn latent correlations between multivariate time-series automatically. In this way, the model emphasizes task-specific correlations in a data-driven fashion.

The multivariate time series $X_{snsr}$ is first fed into a GRU layer, which calculates the hidden state corresponding to each time step $t$ sequentially. Then, we use the last hidden state $\textup{h}$ as the representation of the entire time-series and calculate the weighted adjacent matrix $\mathcal{A}_{snsr}$ by the self-attention mechanism. An attention function can be described as mapping a query and a set of key-value pairs to an output \cite{vaswani2017attention}. The output is computed as a weighted sum of the values, where the weight assigned to each value is computed by a compatibility function of the query with the corresponding key as follows,
\begin{equation}
\begin{split}
    \textup{Query}=\textup{h}\mathcal{W}_{lg}^Q, \; \textup{Key}=\textup{h}\mathcal{W}_{lg}^K\\
    \mathcal{A}_{snsr} = \textup{Softmax}\left(\frac{\textup{Query} \cdot \textup{Key}^T}{\sqrt{d_K}}\right)
\end{split}
\end{equation}
where Query and Key is calculated by linear projections with learnable weights $\mathcal{W}_{lg}^Q$ and $\mathcal{W}_{lg}^K$ in the attention mechanism, respectively; and $d_K$ is the hidden dimension size of $\textup{Key}$. The output matrix $\mathcal{A}_{snsr}\in \mathbb{R}^{N\times N}$ is served as the adjacency weight matrix.

\subsection{Spectral Graph Convolution Module}
After obtaining the graph structured latent representation $\mathcal{G}(X_{snsr},\mathcal{A}_{snsr})$ of the input multivariate time series, the graph $\mathcal{G}$ will be processed by a spectral graph convolution module, as shown in Fig. \ref{fig:spectralGCNblock}. This module is designed to model structural and temporal dependencies jointly in the spectral domain. 

First, a Graph Fourier Transform (GFT) operator $\mathcal{GF}(\cdot)$ transforms the graph $\mathcal{G}$ into a spectral matrix representation on each individual channel $X_i$ of input data, where the uni-variate time-series for each node becomes linearly independent. Then, the output of GFT is fed into the Discrete Fourier Transform (DFT), 1D convolution, GLU, and Inverse Discrete Fourier Transform (IDFT) in order, aiming to decompose each individual sequence into frequency basis and learn feature representations on them. The DFT operator $\mathcal{DF}(\cdot)$ transforms each uni-variate time-series component into the frequency domain. In the frequency domain, the representation is fed into 1D convolution and GLU sub-layers to capture feature patterns in the frequency domain before transformed back to the time domain through IDFT $\mathcal{DF}^{-1}(\cdot)$. The process can be formulated as,
\begin{equation}
\begin{split}
    H_{snsr} & = \sum_{i}\mathcal{DF}^{-1}\left( \textup{GLU}(\mathcal{DF}(\mathcal{GF}(X_{snsr}^i)))\right)\\ &= \sum_{i}\mathcal{DF}^{-1}\left( \textup{GLU}(\theta^{re}_\tau \hat{X}^{re}_u, \theta^{im}_\tau \hat{X}^{im}_u)\right)\\ &= \sum_{i}\theta^{re}_\tau \hat{X}^{re}_u \odot \sigma\left(\theta^{im}_\tau \hat{X}^{im}_u\right)
\end{split}
\end{equation}
where $\hat{X}^{re}_u$ and $\hat{X}^{im}_u$ are the real part and imaginary part of the output of DFT, which are processed by the same operators with different parameters $\theta_\tau$ in parallel. $\theta^{re}_\tau$ and $\theta^{im}_\tau$ are the convolution kernels. $\odot$ is the Hadamard product and nonlinear sigmoid gate $\sigma(\cdot)$ determines how much information in the current input is closely related to the sequential pattern. 

The output from the spectral graph convolution module is fed into fully-connected layers (FCs) to generate sensor features $z_{snsr}$. The FCs composed of 1 layer normalization \cite{ba2016layer}, 1 LeakyReLU activation layer, 1 dropout layer, and 2 stacked linear layers in order.

\subsection{Leverage Textual Information}
In soft sensing, except the \textit{`hard'} sensor types data (including, radar, multi-spectral, acoustic sensor array, etc), the \textit{`soft'} sensor inputs such as textual reports, and hybrid \textit{`hard/soft'} data such as human-annotated sensor data can be highly useful. It is worth categorizing and exploiting to get richer information to enhance the classifier. In this paper, the \textit{`soft'} sensor inputs refer to the textual information of the multi-stage manufacturing process (i.e., categorical variables), where the order of categorical variables is of importance. Textual information should be represented as a fixed-length vector without losing the semantics of the words \cite{shi2017we,shi2018prior}.

Similarly to other sequence transduction models, we use learnable embedding to convert the input tokens and output tokens to vectors of dimension $d_{embd}$. Then, we use a linear layer to map into a hidden textual information representation $z_{embd}$ as,
\begin{equation}
    z_{embd} = \textup{Embedding}(X_{words})*\mathcal{W}_{embd} + b_{embd}
\end{equation}
where $X_{words}$ is the input tokens of words, $\mathcal{W}_{embd}$ and $b_{embd}$ are the weights and bias of the subsequent linear layer, respectively.

In classification problem, not all feature types are equally contributed to the classification task. In order to prioritize the important feature, as shown in Fig. \ref{fig:ssgnn}, we introduce an attention mechanism to capture the dependencies of sensor features $z_{snsr}$ and textual feature $z_{embd}$. The output feature $z_{att}$ is computed as a weighted sum of the $z_{snsr}$, where the weight assigned to each dimension is computed by a compatibility function of the $z_{embd}$ with the corresponding $z_{snsr}$ as follow,
\begin{equation}
\begin{split}
    z_{att} & =\textup{Attention}(z_{embd}, z_{snsr}, z_{snsr})\\ & = \textup{Softmax}(\frac{z_{embd}z_{snsr}^T}{\sqrt{d}})z_{snsr}
\end{split}
\end{equation}
where $d$ is the dimension of $z_{snsr}$, equals to the dimension of $z_{embd}$. We compute the dot products of the textual feature with all sensor features, divide each by $\sqrt{d}$, and apply a $\textup{Softmax}(\cdot)$ function to obtain the weights on $z_{snsr}$.

\subsection{Label Correlation}\label{section: labcorr}
In this paper, we use graph attention networks (GATs) \cite{velivckovic2017graph} to model the inter dependencies between labels. GATs works by information propagation between nodes based on the correlation matrix $\mathcal{A}_{label}$, which is a flexible way to capture the topological structure in the label space. The correlation matrix is not provided in any standard multi-label time series classification datasets. Here, we construct a directed correlation matrix $\mathcal{A}_{label}$ via mining label's co-occurrence patterns within the data. The label correlation dependency is modeled by the form of conditional probability, i.e., $P(l_j |l_i)$ which denotes the probability of occurrence of label $l_j$ when label $l_i$ appears. It is worth mentioning that $P(l_j |l_i)$ is not equal to $P(l_i|l_j)$. For example, when $l_i$ appears in the sample, $l_j$ will also occur with a high probability. However, in the condition of $l_j$ appearing, $l_i$ may not necessarily occur. In other words, the label correlation matrix is asymmetrical.

To construct $\mathcal{A}_{label}$, firstly, we count the occurrence of label pairs in the training set and get the matrix $M \in \mathbb{R}^{C\times C}$, where $C$ is the number of labels, and $M_{ij}$ denotes the concurring times of $l_i$ and $l_j$. Then, we can define the conditional probability matrix by $p_i =M_i/N_i$, where $N_i$ denotes the occurrence times of $l_i$ in the training set, and $p_{ij} = p(l_j |l_i)$ means the probability of label $l_j$ when label $l_i$ appears.

However, the co-occurrence patterns between one label and the other labels may exhibit a long-tail distribution, where some rare co-occurrences may be noise. Such a correlation matrix will over-fit the training data and thus hurt the generalization capacity. Specifically, a threshold $\tau$ is employed prior to filter noisy edges,
\begin{equation}\label{equ:corr_matrix}
    \mathcal{A}_{label}^{ij} = \left\{\begin{matrix}
0 & p_{ij}<\tau\\ 
1 & p_{ij}\geq \tau
\end{matrix}\right.
\end{equation}
where $\mathcal{A}_{label}$ is the binary correlation matrix.

\begin{figure}
    \centering
    \includegraphics[width=0.8\linewidth]{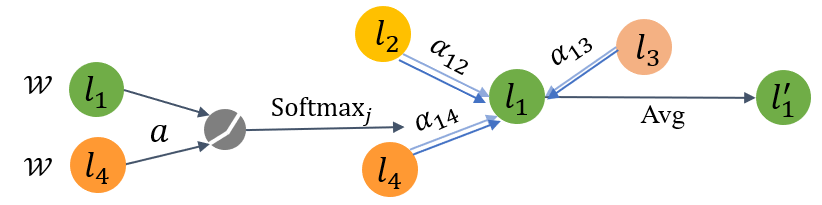}
    \caption{An illustration of multi-head attention (with 2 heads) by node 1 on its 3 neighborhood.}
    \label{fig:lc}
\end{figure}

The motivation of using GATs to model the inter dependencies between labels is by computing the score of each label $l_i$ (node in graph) by attending over its co-occurrence label (neighbors) following a self-attention strategy, as shown in Fig. \ref{fig:lc}. we perform self-attention on the nodes as,
\begin{equation}
    e_{ij}=a(\mathcal{W}l_i, \mathcal{W}l_j)
\end{equation}
where $\mathcal{W}$ is a weight matrix; $a$ is a shared attention mechanism $a:\mathbb{R}\times \mathbb{R}\rightarrow\mathbb{R}$ for computing attention coefficients $e$. $e_{ij}$ indicates the importance of node $i$ to node $j$. We inject the graph structure into the mechanism by performing masked attention, i.e., only compute $e_{ij}$ for nodes $j \in \mathcal{N}_i$, where $\mathcal{N}_i$ is some neighborhood of node $i$ in the graph.

To make coefficients easily comparable across different nodes, softmax function is applied across all choices of $j$. Thus, the normalized coefficients can be obtained by,
\begin{equation}
    \alpha_{ij}=\textup{Softmax}(e_{ij})=\frac{\textup{exp}(e_{ij})}{\sum_{k\in\mathcal{N}_i}\textup{exp}(e_{ik})}
\end{equation}

Then, the normalized attention coefficients $\alpha$ are used to compute a linear combination of the labels scores corresponding to them, to serve as the final output score $l^{'}$ for every node,
\begin{equation}
     l_i^{'}=\sigma\left(\frac{1}{K}\sum_{k=1}^{K} \sum_{j\in\mathcal{N}_i}\alpha_{ij}^k\mathcal{W}^k l_j\right)
\end{equation}
where $K$ indicates the number of independent attention mechanisms, i.e. multi-head attention. $\alpha_{ij}^k$ are normalized attention coefficients computed by the $k$-th attention mechanism ($a^k$), and $\mathcal{W}^k$ is the corresponding input linear transformation's weight matrix. We employ averaging, and delay applying the final nonlinearity activation $\sigma$. 

\subsection{Imbalanced Loss Function}
In a typical multi-label setting, a sample may contain on average few positive labels, and many negative ones. This positive-negative imbalance dominates the optimization process, and can lead to under-emphasizing gradients from positive labels during training, resulting in poor accuracy. Here, we reduce the problem to a series of binary classification tasks. Given $K$ labels, the base network outputs one logit per label, $p_l^k$. Each logit is independently activated by a sigmoid function. Let's denote $y^k_l$ as the ground-truth for class $k$. The total supervised classification loss, $\mathcal{L}_l$, is obtained by aggregating a binary loss from $K$ labels,
\begin{equation}\label{eq:loss1}
\begin{split}
    \mathcal{L}_l = \sum_{k=1}^K & -y_l^k *\mathcal{L}_{pos}^k - (1-y_l^k)*\mathcal{L}_{neg}^k \\ 
    &\mathcal{L}_{pos}^k = -(1-p_l^k)^{\gamma_{+}} \textup{log}(p_l^k) \\ 
    &\mathcal{L}_{neg}^k = -(p_l^k)^{\gamma_{-}}\textup{log}(1-p_l^k)
\end{split}
\end{equation}
where $y_l$ is the ground-truth label, and $\mathcal{L}_{pos}$ and $\mathcal{L}_{neg}$ are the positive and negative focal loss parts (following \cite{lin2017focal}), respectively. $p_l = \sigma(z)$ is the network's output probability and $\gamma$ is the focusing parameter for inner trade-off. $\gamma_{+} =\gamma_{-}= 0$ yields binary cross-entropy. By setting $\gamma > 0$, the contribution of easy negatives (having low probability, $p \ll 0.5$) can be down-weighted in the loss function, enabling to focus more on harder samples during training. Instead of using uniform $\gamma$, we decouple the focusing levels of the positive and negative samples by employing $\gamma_{+}$ and $\gamma_{-}$ be the positive and negative focusing parameters, respectively.

However, we found that simple linear weighting is insufficient to tackle the negative-positive imbalance issue in our case. Instead, following the Asymmetric loss proposed in \cite{ben2020asymmetric}, we use an asymmetric focusing mechanism -- probability shifting -- to perform hard thresholding of very easy negative samples. Which means the negative samples will be fully discarded if their probability is very low. The asymmetric probability-shifted focal loss is defined as,
\begin{equation}\label{Eq:loss2}
    \mathcal{L}_{neg} = (max(p-m, 0))^{\gamma_{-}}\textup{log}(1-max(p-m, 0))
\end{equation}
where $max(p-m, 0)$ is the shifted probability, i.e., moving the loss function to the right by a factor $m$, where $L_{neg}=0$ if $p<m$.

Another major concern in multi-label classification is the high unlabeled rate. Semi-supervised learning provides an effective means of leveraging unlabeled data to improve a classifier's performance. This domain has witnessed rapid progress recently, at the cost of requiring more complexity in models \cite{sohn2020fixmatch}. Inspired by FixMatch proposed in \cite{sohn2020fixmatch}, we introduce a semi-supervised loss term calculated on unlabeled samples by pseudo-labeling method, which uses the model's prediction as a `\textit{label}' to train against. Pseudo-labeling leverages the idea of using the model itself to obtain artificial labels for unlabeled data \cite{scudder1965probability}. Specifically, this refers to leveraging ``hard" labels (i.e., the $\textup{arg max}$ of the model's output) and only retaining artificial labels whose largest class probability fall above a predefined threshold $\varsigma$. Let $p_u$ denotes the model's output of unlabeled sample, $\textup{H}$ denotes the cross-entropy between two probability distributions, $\mathbb{I}$ be the mask, then, the loss term of unlabeled samples is defined as follow, 
\begin{equation}\label{equ:unlabel}
    \mathcal{L}_u =\sum_{k=1}^K\mathbb{I}(p_u^k>\varsigma)\textup{H}(\hat{p}_u^k, p_u^k)
\end{equation}
where $\hat{p}_u^k = \textup{arg max}(p_u^k)$ and $\varsigma$ is the threshold. We assume that $\textup{arg max}$ applied to a probability distribution produces a valid “one-hot” probability distribution.

Finally, the overall loss function is defined as the sum of supervised loss $\mathcal{L}_l$ and unlabeled loss $\mathcal{L}_u$.

\section{Experiment}
\subsection{Seagate Soft-sensing Data} \label{section:data}
The data set is provided by the Seagate manufacturing factories in both Minnesota and Ireland, containing high dimensional time-series sensor data coming from different manufacturing machines. The textual information is the process-relevant categorical variables corresponding to the time-series data. The dataset is publicly available at \url{https://github.com/Seagate/softsensing\_data}.

As shown in Fig. \ref{fig_wafer}, an AlTiC wafer goes through multiple processing stages including polishing, deposition, lithography and etching. After each processing stage, the wafer is sent to metrology tools for quality control inspection, i.e., KQIs. Each metrology stage usually contains a few different measurements, and the same measurement may be performed in different stages. Given there are many-to-many mapping between processes and inspections in each stage, one sensor record are mapped to several KQIs. Each KQI contains a few numerical values to indicate the quality condition of the processed wafer, and a decision of pass/fail is made based on these numbers by engineers at Seagate. For the sake of simplicity, we only cover the pass/fail binary information for each KQI. So that each sample of time-series sensor data are mapped to several binary labels, resulting in a simplified multi-label classification problem. The statistics of the dataset is summarized in TABLE \ref{Tab: dataset}. There are 11 KQIs (labels), and about 1.2\% of them are positive (failed) samples. There are total 194k data samples for training, 34k samples for validation, and 27k samples for testing. The unlabeled rate of training dataset among each labels is list in TABLE \ref{Tab: dataset} column 2. Zero padding is employed in data pre-processing to ensure each sample has fixed 2 time steps.
\begin{table}[t]
\caption{IEEE Big Data Challenge: Soft Sensing at Scale - Seagate Dataset statistics}
\renewcommand{\arraystretch}{1.1}
\begin{adjustbox}{max width=\columnwidth}
\begin{threeparttable}
\begin{tabular}{lcllllll}
\toprule[1.5pt]\midrule[0.5pt]
\multicolumn{1}{c}{\multirow{2}{*}{\begin{tabular}[c]{@{}c@{}}Labels \\ (KQIs)\end{tabular}}}& \multicolumn{1}{c}{\multirow{2}{*}{\begin{tabular}[c]{@{}c@{}}Unlabeled \\ Rate\end{tabular}}} & \multicolumn{2}{c}{Train} & \multicolumn{2}{c}{Valid} & \multicolumn{2}{c}{Test} \\ \cline{3-8} 
\multicolumn{1}{c}{}&\multicolumn{1}{c}{} & \multicolumn{1}{c}{Neg} & \multicolumn{1}{c}{Pos} & \multicolumn{1}{c}{Neg} & \multicolumn{1}{c}{Pos} & \multicolumn{1}{c}{Neg} & \multicolumn{1}{c}{Pos} \\ \midrule[0.5pt]
KQI-1 &0.97 & 6020 & 272 & 1417 & 13 & 878 & 10\\
KQI-2 &0.95 & 10288 & 33 & 1509 & 5 & 950 & 2\\
KQI-3 &0.78 & 42989 & 200 & 7795 & 43 & 5414 & 48\\
KQI-4 &0.94 & 11114 & 132 & 1594 & 23 & 1989 & 33\\
KQI-5 &0.83 & 32794 & 428 & 4283 & 91 & 3567 & 49\\
KQI-6 &0.67 & 64007 & 709 & 11833 & 68 &  9123 & 86\\
KQI-7 &0.39 & 117332 & 1702 & 19663 & 482 & 16975 & 371\\
KQI-8 &0.99 & 1748 & 443 & 196 & 39 & 975 & 8\\
KQI-9 &0.88 & 22420 & 86 & 4225 & 6 & 2906 & 12\\
KQI-10 &0.96 & 7874 & 48 & 1788 & 4 & 1151 & 5\\
KQI-11 &0.81 & 35874 & 227 & 6231 & 36 & 5114 & 43\\  \midrule[0.5pt]\bottomrule[1.5pt]
\end{tabular}
\begin{tablenotes}
\footnotesize
\item \textsuperscript{*} Neg/Pos: the passed/failed samples of corresponding key indicator.
\end{tablenotes}
\end{threeparttable}
\label{Tab: dataset}
\end{adjustbox}
\end{table}

\subsection{Baselines}
\begin{enumerate}
    \item Diagnose recurrent neural network (LSTM) \cite{lipton2015learning} is the first study to empirically evaluate the ability of LSTMs to recognize patterns in multivariate time series of clinical measurements. 
    \item ML-GCN \cite{chen2019multi} is a multi-label classification model based on Graph Convolutional Network (GCN) that is desirable to model the label dependencies to improve the recognition performance. Here, we use two stacked convolution layers as feature extraction backbone and use categorical variables embedding as label representations. The label correlation matrix is set as described in Section \ref{section: labcorr} with $\tau=0.4$.
\end{enumerate}

\subsection{Evaluation Metric}
In order to evaluate our model comprehensively and for the convenience of comparison with other solutions, we report the average per-label, recall (L-R), area under receiver operating curve (L-AUC), the average overall recall (O-R), overall false alarm ratio (O-F), overall AUC (O-AUC) to estimate their effectiveness. For time series sample, the labels are predicted as positive if the confidences of them are greater than 0.5.

\subsection{Implementation Details}
All the baselines and proposed \text{\modelname} are trained on AWS p3.2x large instance with 16 GB NVIDIA Tesla V100 GPU. We implement the models based on PyTorch. Unless otherwise stated, we set $\tau=0.4$ for the correlation matrix in Eq. (\ref{equ:corr_matrix}), set $\varsigma=0.95$ in Eq.(\ref{equ:unlabel}); we adopt LeakyReLU \cite{maas2013rectifier} with the negative slope of $0.2$ as the non-linear activation function. The dropout rate is $0.2$. The dimension is $16$ for embedding, $64$ for $z_{snsr}$ and $z_{embd}$, which is chosen from a search space of [8, 16, 32, 64, 128] on the validation data. For network optimization, RMSProp \cite{ruder2016overview} is used as the optimizer with $1e-4$ weight decay and $1e-3$ initial learning rate. The early stopping mechanism will be executed if the performance of the model on the validation dataset starts to degrade (with patience equals to $25$ epochs).

\section{Results and Discussion}
Quantitative results are reported in TABLE \ref{tbl:nxd}. We compare with state-of-the-art methods, including LSTM, ML-GCN multi-label classification tasks. Here, the L-AUC score for each label is used as the evaluation metric. Here we observe that \text{\modelname} model outperforms the baseline models in most of the labels prediction (highlighted in TABLE \ref{tbl:nxd}), which shows the superiority of our proposed model. By comparing L-AUC scores, \text{\modelname} consistently outperforms LSTM. The major reason lies in that LSTM only takes temporal information into consideration and performs modeling in the time domain, while \text{\modelname} models the time-series data in the frequency domain and shows stable improvement over LSTM.

It is noteworthy that, both \text{\modelname} and ML-GCN that consider label correlation outperforms the LSTMs. It validates the effectiveness of using graphs to model the inter dependencies between labels to improve the classification performance. In both methods, a directed graph is built over labels representations where each node denotes a label, which is a flexible way to capture the topological structure in the label space. Furthermore, \text{\modelname} outperforms ML-GCN in 8 out of 11 KQIs classification. It shows the advantages of leveraging GFT to capture structural information in a graph combined with leveraging DFT to learn temporal patterns, when ML-GCN only use convolutional kernels to extract features.

\begin{table}[!t]
\renewcommand{\arraystretch}{1.05}
\caption{AUC score of \text{\modelname} and Baselines on test dataset}
\label{tbl:nxd}
\centering
\begin{tabular}{llll}
\toprule[1.5pt]\midrule[0.5pt] Labels & \text{\modelname} & ML-GCN & LSTM \\
\midrule
KQI-1 & \textbf{0.76}($\pm$0.027) & 0.48($\pm$0.083) &0.61($\pm$0.007) \\
KQI-2 & 0.57($\pm$0.056) &\textbf{0.60}($\pm$0.088) &0.43($\pm$0.056) \\
KQI-3 & \textbf{0.78}($\pm$0.074) &0.65($\pm$0.031) &0.48($\pm$0.037) \\
KQI-4 & \textbf{0.86}($\pm$0.001) &0.39($\pm$0.065)  &0.49($\pm$0.006) \\
KQI-5 & 0.59($\pm$0.036) &\textbf{0.63}($\pm$0.021)  &0.44($\pm$0.024) \\
KQI-6 & 0.61($\pm$0.025) &\textbf{0.83}($\pm$0.019) &0.53($\pm$0.028) \\
KQI-7 & \textbf{0.67}($\pm$0.003) &0.58($\pm$0.014) &0.51($\pm$0.035) \\
KQI-8 & \textbf{0.78}($\pm$0.057) &0.70($\pm$0.039) &0.38($\pm$0.024) \\
KQI-9 & \textbf{0.87}($\pm$0.029) &0.84($\pm$0.025)  &0.63($\pm$0.009) \\
KQI-10 & \textbf{0.90}($\pm$0.032) &0.58($\pm$0.196)  &0.46($\pm$0.009) \\
KQI-11 & \textbf{0.84}($\pm$0.045) &0.71($\pm$0.032)  &0.62($\pm$0.056) \\
\midrule[0.5pt]\bottomrule[1.5pt]
\end{tabular}
\end{table}

\section{Ablation Study}
In this section, we perform ablation studies from four different aspects, including the advantage of using textual information, effects of label correlation, effects of different loss functions in $\mathcal{L}_l$ for imbalanced multi-label classification, and effects of unlabeled data (with/without $\mathcal{L}_u$).

\subsection{Textual Information - Categorical Variables}
We illustrate that leveraging relevant textual data sources have the potential to improve multi-label classification performance in two ways, data distribution analysis and model performance. To visualize the data distribution, the Isomap \cite{balasubramanian2002isomap} is applied, which is proposed for computing a quasi-isometric, low-dimensional embedding of a set of high-dimensional data points. As shown in Fig.\ref{fig:data_distri}, in the embedded 2D observation space, although the categorical variables result in the fewest feature points, the classifier should not only depend on categorical variables since it only describes how a wafer goes through multiple processing stages. There is no causal relationship between the categorical variables and measurement outcome. As supplementary information, by leveraging categorical variables, the merged data is more `\textit{classifiable}' than sensor data only. To further quantify the effectiveness, we did the ablation study of proposed Soft-sensing GNN w/o categorical variables. As shown in TABLE \ref{tbl:wocat}, we can clearly observe that \text{\modelname} with categorical variables are obviously better than that without categorical variables. Hybrid textual information can be highly useful in Soft sensing.

\begin{figure}
\centering
\includegraphics[width=\linewidth]{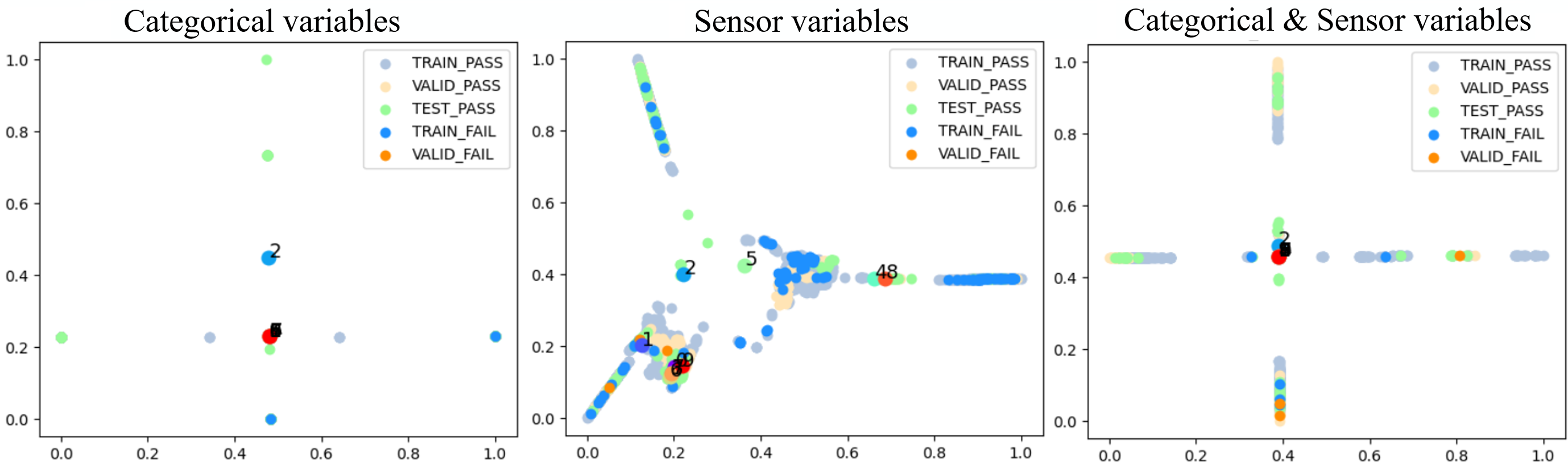}
\caption{Exemplarily data visualization by Isomap embedding method of label KQI-1 by categorical variables only, sensor variables only and both.}
\label{fig:data_distri}
\end{figure}

\begin{table}[!t]
\renewcommand{\arraystretch}{1.05}
\caption{Effectiveness of categorical variables for \text{\modelname}}
\label{tbl:wocat}
\centering
\begin{tabular}{lllll}
\toprule[1.5pt]\midrule[0.5pt] model & Cat vars & O-R & O-F & O-AUC  \\
\midrule
\multirow{2}{*}{\text{\modelname}}& w/ &\textbf{0.51}($\pm 0.043$) &\textbf{0.14}($\pm0.009$) & \textbf{0.75}($\pm 0.016$)\\
& w/o &0.31($\pm 0.129$) &0.24($\pm 0.097$) &0.56($\pm 0.054$)   \\
\midrule[0.5pt]\bottomrule[1.5pt]
\end{tabular}
\end{table}

\subsection{Label Correlation}
A Naive way to address the multi-label classification problem is to treat the labels in isolation, which means convert the multi-label problem into a set of binary classification problems to predict whether a label presents or not. However, this way ignores the topology structure between labels. Especially, in our case, there's some causal relationship between labels. To explore the effectiveness of using GCN to propagate information between multiple labels and consequently learn the inter-dependent relationship for each of the labels, we did the ablation study of proposed \text{\modelname} w/ or w/o label correlation. As shown in TABLE \ref{tbl:wocorr}, it is apparent that our proposed \text{\modelname} with label correlation observes improvements upon the one without label correlation. It approves that capturing the correlations between labels and modeling these label correlations to improve the classifier's performance are both important for multi-label classification.

\begin{table}[!t]
\renewcommand{\arraystretch}{1.05}
\caption{Effectiveness of label correlation for \text{\modelname}}
\label{tbl:wocorr}
\centering
\begin{tabular}{lllll}
\toprule[1.5pt]\midrule[0.5pt] model & Label Corr & O-R & O-F & O-AUC  \\
\midrule
\multirow{2}{*}{\text{\modelname}} & w/ &\textbf{0.51}($\pm 0.043$) &\textbf{0.14}($\pm0.009$) & \textbf{0.75}($\pm 0.016$)\\
& w/o & 0.42($\pm 0.055$) &0.16($\pm 0.030$) & 0.71($\pm 0.042$) \\
\midrule[0.5pt]\bottomrule[1.5pt]
\end{tabular}
\end{table}

\subsection{Loss Function term $\mathcal{L}_l$ for Imbalance}
A key characteristic of multi-label classification is the inherent positive-negative imbalance. Most samples contain only a small fraction of the possible labels, implying that the number of positive samples per category will be, on average, much lower than that of negative samples. In our case, according to TABLE \ref{Tab: dataset}, the average imbalance ratio of $\textup{Neg}/\textup{Pos}=80.02$ for training data (valid:$85.63$, test:$74.73$), and the label \textit{KQI-2}, \textit{KQI-3} and \textit{KQI-9} has top 3 highest high negative-positive imbalance. Another key characteristic of our case study is mislabeling, which could be caused by 2 possible reasons: 1) the dataset collected from both the US and Ireland factories, a wafer could be mislabeled when scrutinized its quality due to consensus conflict which may arise across global engineering teams. 2) the binary label indicating pass/fail is hard encoded based on an internal heuristic threshold value, there exists the possibility of inherent corruption associated with reliance on the threshold.

In this section, we explore how various state-of-the-art loss functions in the supervised $\mathcal{L}_l$ term affect the model performance. Those loss functions are designed for statistically handling the imbalance and mislabeling in multi-label classification problem. By setting $\gamma_+=\gamma_- >0$ in Eq.(\ref{eq:loss1}), we can get the format of Focal loss. Focal loss \cite{lin2017focal} is a common solution to deal with the imbalance in object detection. It puts focus on hard samples, while down-weighting easy samples by decaying the loss as the label’s confidence increases. By setting $\gamma_+=0$ in Eq.(\ref{eq:loss1}) and $\gamma_- >0$ in Eq.(\ref{Eq:loss2}), we can get the format of Asymmetric loss. Asymmetric loss enables \cite{ben2020asymmetric} us to dynamically down-weight and hard-threshold easy negative samples, while also discarding possibly mislabeled samples. 

Here, we compare three different loss functions used in $\mathcal{L}_l$: Binary Cross-entropy, Focal loss and Asymmetric loss. As shown in TABLE \ref{tbl:loss}, the focal loss achieves the best overall recall and AUC while the cross-entropy loss achieves the lowest false positive rate. For cross-entropy loss, 6 out of 11 labels have the lowest false positive rates and highest AUC scores. Furthermore, focal loss and asymmetric loss significantly outperform cross-entropy on this case, demonstrating the effectiveness of $\gamma$ in balancing between negative and positive samples. However, in terms of the recall, which is more important in industrial quality review, we recommend using focal loss training in our proposed \text{\modelname}.

\begin{table}[]
\renewcommand{\arraystretch}{1.1}
\caption{Various $\mathcal{L}_l$ for negative-positive imbalance (per-label)}
\label{tbl:loss}
\centering
\begin{adjustbox}{width=\columnwidth,center}
\begin{threeparttable}
\begin{tabular}{llllllllll}
\toprule[1.5pt]\midrule[0.5pt]
\multicolumn{1}{l}{\multirow{2}{*}{Labels}} & \multicolumn{3}{c}{Cross-entropy} & \multicolumn{3}{c}{Focal loss\cite{lin2017focal}} & \multicolumn{3}{c}{Asymmetric \cite{ben2020asymmetric}} \\\cline{2-10} 
 & L-R & L-F & L-AUC & L-R & L-F & L-AUC & L-R & L-F & L-AUC \\\midrule
KPI-1   &0.00 & \textbf{0.000} &0.73 &0.00 & 0.022 &0.76 &0.00 & 0.012 &\textbf{0.79}\\
KPI-2   &\textbf{0.50} &0.118 &\textbf{0.65} &\textbf{0.50} &0.098 &0.57 &\textbf{0.50} &\textbf{0.095} &0.58\\
KPI-3   &0.54 &\textbf{0.049} &\textbf{0.88} &\textbf{0.67} &0.171 &0.78 &\textbf{0.67} &0.143 &0.79\\
KPI-4   &\textbf{0.61} &0.119 &0.87 &0.45 &0.123 &0.86 &0.42 &\textbf{0.098} &\textbf{0.88}\\
KPI-5   &0.25 &0.280 &0.50 &0.39 &\textbf{0.187} &\textbf{0.59} &\textbf{0.41} &0.231 &0.56\\
KPI-6   &0.34 &\textbf{0.082} &\textbf{0.62} &\textbf{0.36} &0.127 &0.61 &0.34 &0.179 &0.55\\
KPI-7   &0.32 &\textbf{0.141} &\textbf{0.75} &0.53 &0.364 &0.67 &\textbf{0.74} &0.509 &0.67\\
KPI-8   &\textbf{0.56} &0.099 &\textbf{0.92} &0.50 &0.105 &0.78 &0.25 &\textbf{0.082} &0.67\\
KPI-9   &0.33 &\textbf{0.059} &0.58 &\textbf{0.68} &0.129 &\textbf{0.87} &0.36 &0.078 &0.72\\
KPI-10  &0.00 &0.000 &\textbf{0.99} &\textbf{0.80} &0.027 &0.90 &0.40 &\textbf{0.004} &0.84\\
KPI-11  &0.59 &\textbf{0.160} &0.84 &\textbf{0.73} &0.223 &0.84 &0.86 &0.233 &\textbf{0.88}\\\midrule[0.5pt]
\textit{Avg} &0.37 &\textbf{0.101} &\textbf{0.75} &\textbf{0.51} &0.144 &\textbf{0.75} &0.45 &0.151 &0.72 \\
\midrule[0.5pt]\bottomrule[1.5pt]
\end{tabular}
\begin{tablenotes}
\footnotesize
\item \textsuperscript{*} Cross-entropy: $\gamma=0$; Focal loss: $\gamma_+=\gamma_-=2$; Asymmetric loss: $\gamma_+=0, \gamma_-=2$.
\end{tablenotes}
\end{threeparttable}
\end{adjustbox}
\end{table}

\begin{table}[!t]
\renewcommand{\arraystretch}{1.05}
\caption{Effectiveness of unlabeled data for \text{\modelname}}
\label{tbl:wolu}
\centering
\begin{threeparttable}
\begin{tabular}{lllll}
\toprule[1.5pt]\midrule[0.5pt] model & $\mathcal{L}_u$ & O-R & O-F & O-AUC  \\
\midrule
\multirow{2}{*}{\text{\modelname}}& w/ &\textbf{0.51}($\pm 0.043$) &\textbf{0.14}($\pm0.009$) & \textbf{0.75}($\pm 0.016$)\\
& w/o &0.41($\pm 0.051$) &0.14($\pm 0.018$) &0.71($\pm 0.042$)   \\
\midrule[0.5pt]\bottomrule[1.5pt]
\end{tabular}
\begin{tablenotes}
\footnotesize
\item \textsuperscript{*} $\mathcal{L}_l$ Focal loss: $\gamma_+=\gamma_-=2$.
\end{tablenotes}
\end{threeparttable}
\end{table}

\subsection{Leveraging Unlabeled data}
Data-driven models usually achieve their strong performance through supervised learning, which requires a labeled dataset. However, as shown in TABLE \ref{Tab: dataset}, the unlabeled rate (Avg: $0.83$) is extremely high in our case. It's worth leveraging unlabeled data to improve model performance. In this paper, we generate pseudo-labels using the model’s predictions on unlabeled samples. The pseudo-label is only retained if the model assigns a high probability to one of the possible labels, which is realized by using the loss term $\mathcal{L}_u$, shown in Eq.(\ref{equ:unlabel}). TABLE \ref{tbl:wolu} shows that, by adding $\mathcal{L}_u$, \text{\modelname} obtains substantially better performance in terms of O-R (24\% increase) and O-AUC (5.6\% increase). Although the O-F remains the same, the $std$ is halved, indicating a more stable performance.

\section{Conclusion}
This paper proposes a novel graph based soft-sensing neural network (\text{\modelname}) for multivariate time-series classification based on the case of wafer inspection challenge, where the data is noised, highly imbalanced, and unlabeled. In \text{\modelname}, a spectral graph convolution module is introduced to capture the `\textit{classifiable}' intra-series temporal patterns and inter-series sensor correlations jointly in the spectral domain through discrete Fourier transform and graph Fourier transform. Through attention mechanism, the model can leverage both textual information and numerical time series data. In the end, a graph attention network is attached to learn inter-dependent labels prior label representations. Furthermore, we introduce semi-supervised learning based on pseudo-labeling to mitigate the requirement for labeled data by providing a simplified means of leveraging unlabeled data. We also investigate the effectiveness of various loss functions in balancing contributions between negative and positive samples of multi-label classification. Both quantitative and qualitative results validated the advantages of our \text{\modelname} for soft sensor modeling in actual industrial processes.

\section*{Acknowledgment}
This research is partially supported by U.S. National Science Foundation under Grant Nos. IIS-1763452, CNS-1828181, and IIS-2027339. We sincerely thank Seagate Technology for the support on this study, the Seagate Lyve Cloud team for providing the data infrastructure, and the Seagate Open Source Program Office for open sourcing the data sets and the code. Special thanks to the Seagate Data Analytics and Reporting Systems team for inspiring the discussions.

\bibliographystyle{IEEEtran}
\bibliography{IEEEabrv,mybibfile}


\end{document}